\documentclass[11pt,a4paper]{article}
\usepackage[hyperref]{acl2020}
\usepackage{times}
\usepackage{latexsym}
\usepackage{bm}

\usepackage{multirow,multicol} 
\usepackage{booktabs}
\usepackage{mathtools} 
\usepackage{breqn}
\usepackage{float}
\restylefloat{table}
\usepackage{tabularx}
\usepackage{graphics}
\usepackage{adjustbox} 
\usepackage{subfigure}

\usepackage{microtype}

\aclfinalcopy 
\def\aclpaperid{725} 

\newcommand{\PermAcc}{Permutation Acceptance} 
\def\Snospace~{\S{}} 

\newcommand\boldred[1]{\textcolor{red}{\textbf{#1}}}

\DeclareMathOperator*{\argminB}{argmin}

\title{UnNatural Language Inference}

\author{Koustuv Sinha\textsuperscript{1,2,3},
  Prasanna Parthasarathi\textsuperscript{1,2},
  Joelle Pineau\textsuperscript{1,2,3} and
  Adina Williams\textsuperscript{3} \\
  \textsuperscript{1} School of Computer Science, McGill University, Canada \\
  \textsuperscript{2} Montreal Institute of Learning Algorithms (Mila), Canada \\
  \textsuperscript{3} Facebook AI Research (FAIR)\\
  \{koustuv.sinha, prasanna.parthasarathi, jpineau, adinawilliams\}\\@\{mail.mcgill.ca, mail.mcgill.ca, cs.mcgill.ca, fb.com\}}

\date{}

\begin{document}
\maketitle
\begin{abstract}
Recent investigations into the inner-workings of state-of-the-art large-scale pre-trained Transformer-based Natural Language Understanding (NLU) models indicate that they appear to know humanlike syntax, at least to some extent. We provide novel evidence that complicates this claim: we find that state-of-the-art Natural Language Inference (NLI) models assign the same labels to permuted examples as they do to the original, i.e. they are largely invariant to random word-order permutations. This behavior notably differs from that of humans; we struggle with ungrammatical sentences. To measure the severity of this issue, we propose a suite of metrics and investigate which properties of particular permutations lead models to be word-order invariant. In the MNLI dataset, for example, we find almost all (98.7\%) examples contain at least one permutation which elicits the gold label. Models are sometimes even able to assign gold labels to permutations that they originally failed to predict correctly. We provide a comprehensive empirical evaluation of this phenomenon, and further show that this issue exists for both Transformers and pre-Transformer RNN / ConvNet based encoders, as well as across multiple languages (English and Mandarin Chinese). Our code and data are available at \href{https://github.com/facebookresearch/unlu}{https://github.com/facebookresearch/unlu}. 
\end{abstract}

\section{Introduction}

Of late, large scale pre-trained Transformer-based \citep{vaswani-etal-2017-attention} models---such as RoBERTa \citep{liu-et-al-2019-roberta}, BART \citep{lewis-etal-2020-bart}, and GPT-2 and -3 \citep{radford-etal-2019-language,brown-etal-2020-gpt3}---have exceeded recurrent neural networks' performance on many NLU tasks \citep{wang-etal-2018-glue, wang-etal-2019-superglue}. 
Several papers have even suggested that Transformers pretrained on a language modeling (LM) objective can capture syntactic information \citep{hewitt-manning-2019-structural,jawahar-etal-2019-bert, warstadt-bowman-2020-can, wu-etal-2020-perturbed}, with their self-attention layers being capable of surprisingly effective learning \cite{rogers2020}. In this work, we question such claims that current models ``know syntax''. 

\begin{table}[t]
    \centering
    \small
    \begin{adjustbox}{max width=0.95\linewidth}
    \begin{tabular}{p{11em}p{9em}p{3em}} 
    \toprule
     \bf Premise & \bf Hypothesis & \bf Predicted Label \\ \midrule
    Boats in daily use lie within feet of the fashionable bars and restaurants.  & There are boats close to bars and restaurants. & E \\ 
    \addlinespace[0.5em]
    restaurants and use feet of fashionable lie the in Boats within bars daily . & bars restaurants are There and to close boats . & E \\ \midrule
    He and his associates weren't operating at the level of metaphor. & He and his associates were operating at the level of the metaphor. & C\\  \addlinespace[0.5em]
    his at and metaphor the of were He operating associates n't level . & his the and metaphor level the were He at associates operating of . & C\\ 
    \bottomrule
    \end{tabular}
   \end{adjustbox}
    \caption{Examples from the MNLI Matched development set. Both the original example and the permuted one elicit the same classification label (entailment and contradiction respectively) from RoBERTa (large). 
    A simple demo is provided in an associated \href{https://colab.research.google.com/drive/1vv8Xmag1go3dib4vZXUZXAFB4ltDaMH7?usp=sharing}{Google Colab notebook.}}
    \label{tab:example}
\end{table}

Since there are many ways to investigate ``syntax'', we must be clear on what we mean by the term.  
Knowing the syntax of a sentence means being sensitive to the \textit{order of the words} in that sentence (among other things).  Humans are sensitive to word order, so clearly, ``language is not merely a bag of words'' \citep[p.156]{harris-1954-distributional}.
Moreover, it is easier for us to identify or recall words presented in canonical orders than in disordered, ungrammatical sentences; this phenomenon is called the \textit{``sentence superiority effect''} (\citealt{cattell-1886-time, scheerer1981early, toyota-2001-changes, baddeley-etal-2009-working, snell-grainger-2017-sentence, snell2019word, wen-etal-2019-parallel}, i.a.).   
In our estimation then, if one wants to claim that  a model ``knows syntax'', then they should minimally show that the model is sensitive to word order (at least for e.g. English or Mandarin Chinese).

Generally, knowing the syntax of a sentence is taken to be a prerequisite for understanding what that sentence means \citep{heim-kratzer-1998-semantics}.
Models should have to know the syntax first then, if performing any particular NLU task that genuinely requires a humanlike understanding of meaning (cf. \citealt{bender-koller-2020-climbing}). 
Thus, if our models are as good at NLU as our current evaluation methods suggest, we should expect them to be sensitive to word order (see \autoref{tab:example}).  We find, based on a suite of permutation metrics, that they are not.


We focus here on textual entailment, one of the hallmark tasks used to measure how well models understand language \citep{condoravdi-etal-2003-entailment, dagan-etal-2005-pascal}. This task, often also called Natural Language Inference (NLI; \citealt{bowman-etal-2015-large}, i.a.), typically consists of two sentences: a premise and a hypothesis. The objective is to predict whether the premise entails the hypothesis, contradicts it, or is neutral with respect to it.  We find rampant word order insensitivity in purportedly high performing NLI models. 
For nearly all premise-hypothesis pairs, \textbf{there are many permuted examples that fool the models} into providing the correct prediction. In case of MNLI, for example, the current state-of-the-art of 90.5\% can be increased to \textbf{98.7}\% merely by permuting the word order of test set examples. We even find drastically increased cross-dataset generalization when we reorder words. This is not just a matter of chance---we show that the model output probabilities are significantly different from uniform. 

We verify our findings with three popular English NLI datasets---SNLI \citep{bowman-etal-2015-large}, MultiNLI \citep{williams-etal-2018-broad} and ANLI \citep{nie-etal-2020-adversarial})---
and one Chinese one, OCNLI \cite{hu-etal-2020-ocnli}. It is thus less likely that our findings result from some quirk of English or a particular tokenization strategy. 
We also observe the effect for various transformer architectures pre-trained on language modeling (BERT, RoBERTa, DistilBERT), and non-transformers, including a ConvNet, an InferSent model, and a BiLSTM.

Our contributions are as follows: 
(i) we propose a suite of metrics (\textit{\PermAcc}) for measuring model insensitivity to word order (\autoref{sec:permutation}), 
(ii) we construct multiple permuted test datasets for measuring NLI model performance at a large scale (\autoref{sec:eval}),
(iii) we show that NLI models focus on words more than word order, but can partially reconstruct syntactic information from words alone (\autoref{sec:pos_mini_tree}), 
(iv) we show the problem persists on out-of-domain data,
(v) we show that humans struggle with UnNatural Language Inference, underscoring the non-humanlikeness of SOTA models (\autoref{sec:human_eval}),
(vi) finally, we explore a simple maximum entropy-based method (\autoref{sec:training}) to encourage models not to accept permuted examples.


\section{Related Work}

Researchers in NLP have realized the importance of syntactic structure in neural networks going back to \newcite{tabor-1994-syntactic}. An early hand annotation effort on PASCAL RTE \citep{dagan-etal-2006} suggested that ``syntactic information alone was sufficient to make a judgment'' for roughly one third of examples 
\citep{vanderwende2005syntax}. %
Anecdotally, large generative language models like GPT-2 or -3 exhibit a seemingly humanlike ability to generate fluent and grammatical text \citep{goldberg2019assessing, wolf2019some}. 
However, the jury is still out as to whether transformers genuinely acquire syntax.

\paragraph{Models appear to have acquired syntax.} When researchers have peeked inside Transformer LM's pretrained representations, familiar syntactic structure \citep{hewitt-manning-2019-structural,jawahar-etal-2019-bert, lin-etal-2019-open, warstadt-bowman-2020-can, wu-etal-2020-perturbed}, or a familiar order of linguistic operations \citep{jawahar-etal-2019-bert,  tenney-etal-2019-bert}, has appeared. There is also evidence, notably from agreement attraction phenomena \citep{linzen-etal-2016-assessing} that transformer-based models pretrained on LM do acquire some knowledge of natural language syntax \citep{gulordava-etal-2018-colorless, chrupala-alishahi-2019-correlating,  jawahar-etal-2019-bert, lin-etal-2019-open, manning-etal-2020-emergent,  hawkins-etal-2020-investigating, linzen-baroni-2021-syntactic}. Results from other phenomena \citep{warstadt-bowman-2020-can} such as NPI licensing  \citep{warstadt-etal-2019-investigating} lend additional support. The claim that LMs acquire some syntactic knowledge has been made not only for transformers, but also for convolutional neural nets \citep{bernardy-lappin-2017-using}, and RNNs \citep{gulordava-etal-2018-colorless, van-schijndel-linzen-2018-neural, wilcox-etal-2018-rnn, zhang-bowman-2018-language, prasad-etal-2019-using, ravfogel-etal-2019-studying}---although there are many caveats (e.g., \citealt{ravfogel-etal-2018-lstm, white-etal-2018-lexicosyntactic,  davis-van-schijndel-2020-recurrent, chaves-2020-dont, da-costa-chaves-2020-assessing, kodner-gupta-2020-overestimation}). 

\paragraph{Models appear to struggle with syntax.} Several works have cast doubt on the extent to which NLI models in particular know syntax (although each work adopts a slightly different idea of what ``knowing syntax'' entails). For example, \newcite{mccoy-etal-2019-right} argued that the knowledge acquired by models trained on NLI (for at least some popular datasets) is actually not as syntactically sophisticated as it might have initially seemed; some transformer models rely mainly on simpler, non-humanlike heuristics. In general, transformer LM performance has been found to be patchy and variable across linguistic phenomena \citep{dasgupta-etal-2018-evaluating, naik-etal-2018-stress, an-etal-2019-representation, ravichander-etal-2019-equate, jeretic-etal-2020-natural}. This is especially true for syntactic phenomena \citep{marvin-linzen-2018-targeted, hu-etal-2020-systematic, gauthier-etal-2020-syntaxgym, mccoy-etal-2020-berts, warstadt-etal-2020-blimp}, where transformers are, for some phenomena and settings, worse than RNNs \citep{van-schijndel-etal-2019-quantity}. From another angle, many have explored architectural approaches for increasing a network's sensitivity to syntactic structure \citep{chen-etal-2017-enhanced, Li-etal-2020-SANLI}. \newcite{williams-etal-2018-latent} showed that learning jointly to perform NLI  and to parse resulted in parse trees that match no popular syntactic formalisms. Furthermore, models trained explicitly to differentiate acceptable sentences from unacceptable ones (i.e., one of the most common syntactic tests used by linguists) have, to date, come nowhere near human performance \citep{warstadt-etal-2019-neural}.

\paragraph{Insensitivity to Perturbation.} Most relatedly, several concurrent works \citep{pham-etal-2020-out, alleman2021syntactic, gupta-etal-2021-bert, sinha2021masked,parthasarathi2021sometimes} investigated the effect of word order permutations on transformer NNs.  \newcite{pham-etal-2020-out} is very nearly a proper subset of our work except for investigating additional tasks (i.e. from the GLUE benchmark of \citealt{wang-etal-2018-glue}) and performing a by-layer-analysis. \newcite{gupta-etal-2021-bert} also relies on the GLUE benchmark, but additionally investigates other types of ``destructive'' perturbations. Our contribution differs from these works 
in that we additionally include the following: we (i) outline theoretically-informed predictions for how models \emph{should be expected} to react to permuted input (we outline a few options), (ii) show that permuting can ``flip'' an incorrect prediction to a correct one, (iii) show that the problem isn't specific to Transformers, (iv) show that the problem persists on out of domain data, (v) offer a suite of flexible metrics, and (vi) analyze \emph{why} models might be accepting permutations (BLEU and POS-tag neighborhood analysis). Finally, we replicate our findings in another language.
While our work (and \citeauthor{pham-etal-2020-out,gupta-etal-2021-bert}) only permutes data during fine-tuning and/or evaluation, 
recently \citeauthor{sinha2021masked} explored the sensitivity during pre-training, and found that models trained on n-gram permuted sentences perform remarkably close to regular MLM pre-training.
In the context of generation, \newcite{parthasarathi2021sometimes} crafted linguistically relevant perturbations (on the basis of part-of-speech tagging and dependency parsing) to evaluate whether permutation hinders automatic machine translation models. Relatedly, but not for translation, \newcite{alleman2021syntactic} investigated a smaller inventory of perturbations with emphasis on phrasal boundaries and the effects of n-gram perturbations on different layers in the network. 

\paragraph{NLI Models are very sensitive to words.} NLI models often over-attend to particular words to predict the correct answer \citep{gururangan-etal-2018-annotation, clark-etal-2019-bert}. \newcite{wallace-etal-2019-universal} show that some short sequences of non-human-readable text can fool many NLU models, including NLI models trained on SNLI, into predicting a specific label. In fact, \newcite{ettinger-2020-whatbertisnot} observed that for one of three test sets, BERT loses some accuracy in word-perturbed sentences, but that there exists a subset of examples for which BERT’s accuracy remains intact. 
If performance isn't affected (or if permutation helps, as we find it does in some cases), it suggests that these state-of-the-art models actually perform somewhat similarly to bag-of-words models \cite{blei-etal-2003-latent, mikolov2013efficient}.   


\section{Our Approach}\label{sec:permutation}

As we mentioned, linguists generally take syntactic structure to be necessary for humans to know what sentences mean. Many also find the NLI task to a very promising approximation of human natural language understanding, in part because it is rooted in the tradition of logical entailment. In the spirit of propositional logic, sentence meaning is taken to be 
truth-conditional \citep{frege1948sense, montague-1970-universal, chierchia-mcconnell-1990-meaning, heim-kratzer-1998-semantics}. That is to say that understanding a sentence is equivalent to knowing the actual conditions of the world under which the sentences would be (judged) true \citep{wittgenstein-1922-tractatus}. If grammatical sentences are required for sentential inference, as per a truth conditional approach \citep{montague-1970-universal}, then permuted sentences should be meaningless. Put another way, the meanings of highly permuted sentences (if they exist) are not propositions, and thus those sentences don't have truth conditions. Only from their truth conditions of sentences can we tell if a sentence entails another. 
In short, the textual entailment task is technically undefined in our ``unnatural'' setting. 

Since existing definitions don't immediately extend to UnNatural Language Inference (UNLI), we outline several hypothetical \textit{systematic} ways that a model might perform, had it been sensitive to word order. We hypothesize two models that operate on the first principles of NLI, and one that doesn't. In the first case, Model A deems permuted sentences meaningless (devoid of truth values), as formal semantic theories of human language would predict. Thus, it assigns ``neutral" to every permuted example. Next, Model B does not deem permuted sentences meaningless, and attempts to understand them. Humans find understanding permuted sentences difficult (see our human evaluations in \autoref{sec:human_eval}). Model B could also similarly struggle to decipher the meaning, and just equally sample labels for each example (i.e., assigns equal probability mass to the outcome of each label). Finally, we hypothesize a non-systematic model, Model C, which attempts to treat permuted sentences as though they weren't permuted at all. This model could operate similarly as bag-of-words (BOW), and thus always assign the same label to the permuted examples as it would to the un-permuted examples. If the model failed to assign the original gold label to the original unpermuted examples, it will also fail to assign the original gold label to its permutations; it will never get higher accuracy on permuted examples than on unpermuted ones.

We find in our experiments that the state-of-the-art Transformer-based NLI models (as well as pre-Transformer class of models) do not perform like any of the above hypothetical models. They perform closest to Model C, but are, in some cases, actually able to achieve \emph{higher} accuracy on permuted examples. To better quantitatively describe this behaviour,
we introduce our suite of \textbf{\PermAcc} metrics that enable us to quantify how accepting models are of permuted sentences. 

\section{Methods}\label{sec:constructpermut}

\paragraph{Constructing the permuted dataset.} For a given dataset $D$ having splits $D_{\text{train}}$ and $D_{\text{test}}$, we first train an NLI model $M$ on $D_{\text{train}}$ to achieve comparable accuracy to what was reported in the original papers. We then construct a randomized version of $D_{\text{test}}$, which we term as $\hat{D}_{\text{test}}$ such that: for each example $(p_i,h_i,y_i) \in D_{\text{test}}$ (where $p_i$ and $h_i$ are the premise and hypothesis sentences of the example respectively and $y_i$ is the gold label), we use a permutation operator $\mathcal{F}$ that returns a list ($\hat{P}_i, \hat{H}_i$) of $q$ permuted sentences ($\hat{p}_i$ and $\hat{h}_i$), where $q$ is a hyperparameter. $\mathcal{F}$ essentially permutes all positions of the words in a given sentence (i.e., either in premise or hypothesis) with the restriction that \textit{no words maintain their original position}.  In our initial setting, we do not explicitly control the placement of the words relative to their original neighbors, but we analyze clumping effects in \autoref{sec:eval}.
$\hat{D}_{\text{test}}$ now consists of $|D_{\text{test}}| \times q$ examples, with $q$ different permutations of hypothesis and premise for each original test example pair. If a sentence $S$ (e.g., $h_i$) contains $w$ words, then the total number of available permutations of $S$ are $(w-1)!$, thus making the output of $\mathcal{F}$ a list of $(w-1)! \choose q$ permutations in this case. For us, the space of possible outputs is larger, since we permute $p_i$ and $h_i$ separately (and ignore examples for which any $|S|\leq5$).

\paragraph{Defining \PermAcc.}
The choice of $q$ naturally allows us to analyze a statistical view of the predictability of a model on the permuted sentences. To that end, we define the following notational conventions. Let $\mathcal{A}$ be the original accuracy of a given model $M$ on a dataset $D$, and \textit{c} be the number of examples in a dataset which are marked as correct according to the standard formulation of accuracy for the original dataset (i.e., they are assigned the ground truth label). Typically $\mathcal{A}$ is given by $\frac{c}{|D_{test}|}$ or $\frac{c}{|D_{dev}|}$. 

\begin{figure}[t]
    \centering
    \resizebox{0.5\textwidth}{!}{
        \includegraphics{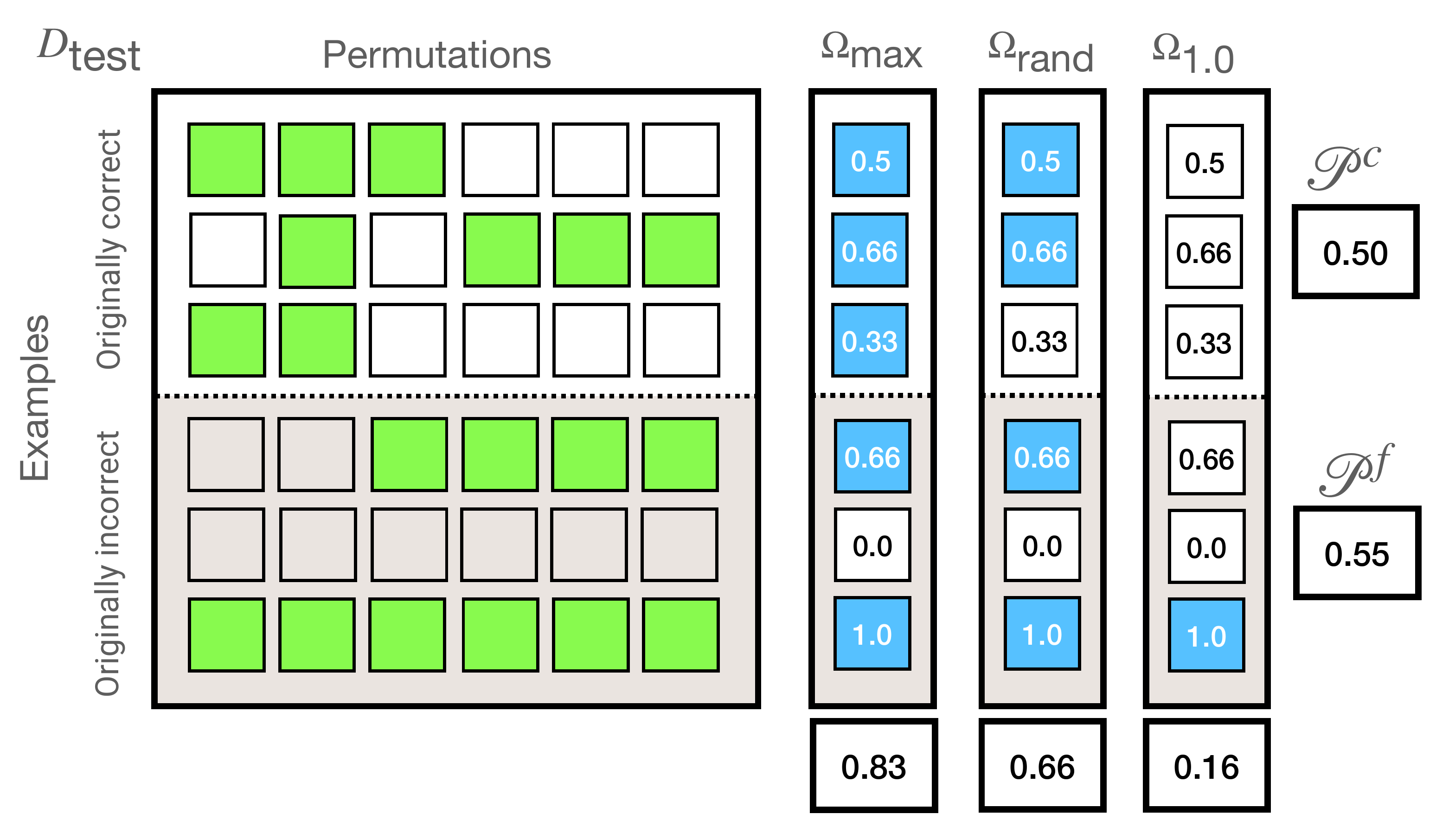}}
    \caption{Graphical representation of the \PermAcc\ class of metrics. Given a sample test set ${D}_{\text{test}}$ with six examples, three of which originally predicted correctly (model predicts gold label), three incorrectly (model fails to predict gold label), with $n=6$ permutations, $\Omega_{\text{max}}$,$\Omega_{\text{rand}}$, $\Omega_{\text{1.0}}$, $\mathcal{P}^c$ and $\mathcal{P}^f$ are provided. Green boxes indicate permutations accepted by the model. Blue boxes mark examples that crossed each threshold and were used to compute the corresponding metric. }
    \label{fig:def_metrics}
\end{figure}

Let $\Pr_{M}(\hat{P}_i, \hat{H}_i)_{\text{cor}}$ then be the percentage of $q$ permutations of an example $(p_i, h_i)$ assigned the ground truth label $y_i$ by $M$: 
    \begin{align}
    &&\Pr_{M}(\hat{P}_i, \hat{H}_i)_{\text{cor}} =  
    \frac{1}{q}\sum_{\mathclap{(\hat{p}_j \in \hat{P}_i, \hat{h}_j \in \hat{H}_i)}}((M(\hat{p}_j, \hat{h}_j) = y_i) \rightarrow 1)
    \end{align}
To get an overall summary score, we let $\Omega_x$ be the percentage of examples $(p_i, h_i) \in D_{\text{test}}$ for which $\Pr_{M}(\hat{P}_i, \hat{H}_i)_{\text{cor}}$ exceeds a predetermined threshold $0 < x < 1$. 
Concretely, a given example will count as correct according to $\Omega_x$ if more than $x$ percent of its permutations ($\hat{P}_i$ and $\hat{H}_i$) are assigned $y_i$ by the model $M$.
Mathematically, 
    \begin{align} 
    &&\Omega_x = \frac{1}{\mid D_{test}\mid} \sum_{\mathrlap{(p_i, h_i)\in D_{test}}} ((\Pr_{M}(\hat{P}_i, \hat{H}_i)_{\text{cor}} 
    > x) \rightarrow 1).
    \end{align}
There are two specific cases of $\Omega_{\text{x}}$ that we are most interested in. First, we define $\Omega_{\text{max}}$ or the \textbf{Maximum Accuracy}, where $x = 1 / |D_{\text{test}}|$. In short, $\Omega_{\text{max}}$ gives the percentage of examples $(p_i, h_i) \in D_{\text{test}}$ for which there is \textit{at least one} permutation $(\hat{p_j}, \hat{h_j})$ that model $M$ assigns the gold label $y_i$
\footnote{Theoretically, $\Omega_{\text{max}} \rightarrow 1$ if the number of permutations $q$ is large. Thus, in our experiments we set $q=100$.}.
Second, we define $\Omega_{\text{rand}}$, or \textbf{Random Baseline Accuracy}, where $x = 1 / m$ or chance probability (for balanced $m$-way classification, where $m=3$ in NLI). This metric is less stringent than $\Omega_{\text{max}}$, as it counts an example if at least \textit{one third} of its permutations are assigned the gold label (hence provides a lower-bound relaxation). See \autoref{fig:def_metrics} for a graphical representation of $\Omega_{\text{x}}$.

We also define 
$D^{f}$ to be the list of examples originally marked incorrect according to $\mathcal{A}$, but are now deemed correct according $\Omega_{\text{max}}$. $D^{c}$ is the list of examples originally marked correct according to  $\mathcal{A}$. Thus, we should expect $D^{f}<D^{c}$ for models that have high accuracy. 
Additionally, we define $\mathcal{P}^c$ and $\mathcal{P}^f$, as the dataset average percentage of permutations which predicted the gold label, when the examples were originally correct ($D^{c}$) and when the examples were originally incorrect ($D^{f}$) as per $\mathcal{A}$ (hence, flipped) respectively.
\begin{equation}
\begin{split}
    &\mathcal{P}^{c} = \frac{1}{|D^{c}|} \sum_{i=0}^{|D^{c}|} M(\hat{P}_i, \hat{H}_i)_{\text{cor}}\\
\end{split}
\end{equation}

\noindent $P^f$ is defined similarly by replacing $D^c$ by $D^f$. Note that for a classic BOW model,  $\mathcal{P}^c=100$ and $\mathcal{P}^f=0$, because it would rely on the words alone (not their order) to make its classification decision. Since permuting removes no words, BOW models should come to the same decisions for permuted examples as for the originals.

\begin{table}[htbp]
  \centering
  \resizebox{\linewidth}{!}{%
    \begin{tabular}{@{}llrrrrr@{}}
\toprule
           Model &    Eval. Dataset  &  $\mathcal{A}$ &  $\Omega_{\text{max}}$ &  $\mathcal{P}^c$ &  $\mathcal{P}^f$ &  $\Omega_{\text{rand}}$ \\
\midrule
\multirow{7}{*}{\bf RoBERTa-Large} 
 &   MNLI\_m\_dev &              0.906 &         0.987 &                  0.707 &             0.383 &                        0.794 \\
 &  MNLI\_mm\_dev &              0.901 &         0.987 &                  0.707 &             0.387 &                        0.790 \\
 &     SNLI\_dev &              0.879 &         0.988 &                  0.768 &             0.393 &                        0.826 \\
 &    SNLI\_test &              0.883 &         0.988 &                  0.760 &             0.407 &                        0.828 \\
 &  A1* &              0.456 &         0.897 &                  0.392 &             0.286 &                        0.364 \\
 &  A2* &              0.271 &         0.889 &                  0.465 &             0.292 &                        0.359 \\
 &  A3* &              0.268 &         0.902 &                  0.480 &             0.308 &                        0.397 \\ \midrule
 & Mean & 0.652 & 0.948 & 0.611 & \boldred{0.351} & 0.623 \\ 
\midrule
 
\multirow{7}{*}{\bf BART-Large} 
    &   MNLI\_m\_dev &              0.902 &         0.989 &                  0.689 &             0.393 &                        0.784 \\
    &  MNLI\_mm\_dev &              0.900 &         0.986 &                  0.695 &             0.399 &                        0.788 \\
    &     SNLI\_dev &              0.886 &         0.991 &                  0.762 &             0.363 &                        0.834 \\
    &    SNLI\_test &              0.888 &         0.990 &                  0.762 &             0.370 &                        0.836 \\
    &  A1* &              0.455 &         0.894 &                  0.379 &             0.295 &                        0.374 \\
    &  A2* &              0.316 &         0.887 &                  0.428 &             0.303 &                        0.397 \\
    &  A3* &              0.327 &         0.931 &                  0.428 &             0.333 &                        0.424 \\ \midrule
& Mean &  \textbf{0.668} & \boldred{0.953} & 0.592 & \boldred{0.351} & \boldred{0.634} \\
\midrule
\multirow{7}{*}{\bf DistilBERT}  &   MNLI\_m\_dev &              0.800 &         0.968 &                  0.775 &             0.343 &                        0.779 \\
      &  MNLI\_mm\_dev &              0.811 &         0.968 &                  0.775 &             0.346 &                        0.786 \\
      &     SNLI\_dev &              0.732 &         0.956 &                  0.767 &             0.307 &                        0.731 \\
      &    SNLI\_test &              0.738 &         0.950 &                  0.770 &             0.312 &                        0.725 \\
      &  A1* &              0.251 &         0.750 &                  0.511 &             0.267 &                        0.300 \\
      &  A2* &              0.300 &         0.760 &                  0.619 &             0.265 &                        0.343 \\
      &  A3* &              0.312 &         0.830 &                  0.559 &             0.259 &                        0.363 \\ \midrule
      & Mean &  0.564 & 0.883 & \boldred{0.682} & 0.300 & 0.575 \\ 
\midrule\midrule
 \multirow{7}{*}{\bf InferSent} 
 &   MNLI\_m\_dev &              0.658 &         0.904 &                  0.842 &             0.359 &                        0.712 \\
 &  MNLI\_mm\_dev &              0.669 &         0.905 &                  0.844 &             0.368 &                        0.723 \\
 &     SNLI\_dev &              0.556 &         0.820 &                  0.821 &             0.323 &                        0.587 \\
 &    SNLI\_test &              0.560 &         0.826 &                  0.824 &             0.321 &                        0.600 \\
 &  A1* &              0.316 &         0.669 &                  0.425 &             0.395 &                        0.313 \\
 &  A2* &              0.310 &         0.662 &                  0.689 &             0.249 &                        0.330 \\
 &  A3* &              0.300 &         0.677 &                  0.675 &             0.236 &                        0.332 \\ \midrule
 & Mean &  \textbf{0.481} & 0.780 & 0.731 & \boldred{0.322} & 0.514 \\ 
 \midrule
 \multirow{7}{*}{\bf ConvNet}
 &   MNLI\_m\_dev &              0.631 &         0.926 &                  0.773 &             0.340 &                        0.684 \\
 &  MNLI\_mm\_dev &              0.640 &         0.926 &                  0.782 &             0.343 &                        0.694 \\
 &     SNLI\_dev &              0.506 &         0.819 &                  0.813 &             0.339 &                        0.597 \\
 &    SNLI\_test &              0.501 &         0.821 &                  0.809 &             0.341 &                        0.596 \\
 &  A1* &              0.271 &         0.708 &                  0.648 &             0.218 &                        0.316 \\
 &  A2* &              0.307 &         0.725 &                  0.703 &             0.224 &                        0.356 \\
 &  A3* &              0.306 &         0.798 &                  0.688 &             0.234 &                        0.388 \\ \midrule
 & Mean &  0.452 & \boldred{0.817} & \boldred{0.745} & 0.291 & 0.519 \\ 
\midrule
 \multirow{7}{*}{\bf BiLSTM} 
 &   MNLI\_m\_dev &              0.662 &         0.925 &                  0.800 &             0.351 &                        0.711 \\
 &  MNLI\_mm\_dev &              0.681 &         0.924 &                  0.809 &             0.344 &                        0.724 \\
 &     SNLI\_dev &              0.547 &         0.860 &                  0.762 &             0.351 &                        0.598 \\
 &    SNLI\_test &              0.552 &         0.862 &                  0.771 &             0.363 &                        0.607 \\
 &  A1* &              0.262 &         0.671 &                  0.648 &             0.271 &                        0.340 \\
 &  A2* &              0.297 &         0.728 &                  0.672 &             0.209 &                        0.328 \\
 &  A3* &              0.304 &         0.731 &                  0.656 &             0.219 &                        0.331 \\ \midrule
 & Mean &  0.472 & 0.814 & 0.731 & 0.301 & \boldred{0.520} \\
 
\bottomrule
\end{tabular}}
  \caption{Statistics for Transformer-based models trained on MNLI corpus \cite{williams-etal-2018-broad}. 
  The highest values are bolded (\boldred{red} indicates the model most insensitive to permutation) per metric and per model class (Transformers and non-Transformers). A1*, A2* and A3* refer to the ANLI dev. sets \citep{nie-etal-2020-adversarial}.}
  \label{table:main}
\end{table}
\begin{table}[htbp]
    \centering
    \footnotesize
    \resizebox{\linewidth}{!}{%
        \begin{tabular}{llrrrrrr}
            \toprule
             Model              & $\mathcal{A}$ & $\Omega_{\text{max}}$ & $\mathcal{P}^c$ & $\mathcal{P}^f$ & $\Omega_{\text{rand}}$ \\ \midrule
             RoBERTa-Large   & \textbf{0.784} &         \boldred{0.988} &                  0.726 &             \boldred{0.339} &                        \boldred{0.773}            \\
             InferSent &             0.573 &         0.931 &                  0.771 &             0.265 &                        0.615 \\
   ConvNet &              0.407 &         0.752 &                  \boldred{0.808} &             0.199 &                        0.426 \\
    BiLSTM &              0.566 &         0.963 &                  0.701 &             0.271 &                        0.611 \\
            \bottomrule
        \end{tabular}}
    \caption{Results on evaluation on OCNLI Dev set. All models are trained on OCNLI corpus \cite{hu-etal-2020-ocnli}. 
    Bold marks the highest value per metric (\boldred{red} shows the model is insensitive to permutation).}
    \label{table:ocnli_all}
\end{table}

\section{Results}\label{sec:eval} 

We present results for two types of models: \textbf{(a)} Transformer-based models and \textbf{(b)} Non-Transformer Models. In \textbf{(a)}, we investigate the state-of-the-art pre-trained models such as RoBERTa-Large \cite{liu-et-al-2019-roberta}, BART-Large \cite{lewis-etal-2020-bart} and DistilBERT \cite{sanh2020distilbert}. For \textbf{(b)} we consider several recurrent and convolution based neural networks, such as InferSent \cite{conneau-etal-2017-supervised}, Bidirectional LSTM \cite{collobert2008unified} and ConvNet \cite{zhao2015self}. We train all models on MNLI, and evaluate on in-distribution (SNLI and MNLI) and out-of-distribution datasets (ANLI). We independently verify results of \textbf{(a)} using both our fine-tuned model using HuggingFace Transformers \cite{wolf2020transformers} and pre-trained checkpoints from FairSeq \cite{ott2019fairseq} (using PyTorch Model Hub). For \textbf{(b)}, we use the InferSent codebase. We sample $q=100$ permutations for each example in $D_{\text{test}}$, and use 100 seeds for each of those permutations to ensure full reproducibility. We drop examples from test sets where we are unable to compute \textit{all unique} randomizations, typically these are examples with sentences of length of less than 6 tokens. \footnote{Code, data, and model checkpoints will be available at \href{https://github.com/facebookresearch/unlu}{https://github.com/facebookresearch/unlu}.}

\paragraph{Models accept many permuted examples.}

We find $\Omega_{\text{max}}$ is very high for models trained and evaluated on MNLI (in-domain generalization), reaching \textbf{98.7\%} on MNLI dev. and test sets (in RoBERTa, compared to $\mathcal{A}$ of 90.6\% (\autoref{table:main}). Recall, human accuracy is approximately 92\% on MNLI dev.,  \citealt{nangia-bowman-2019-human}). This shows that there exists at least one permutation (usually many more) for almost all examples in $D_{\text{test}}$ such that model $M$ predicts the gold label. We also observe high $\Omega_{\text{rand}}$ at 79.4\%, showing that there are many examples for which the models outperform even a random baseline in accepting permuted sentences (see \autoref{app_sec:threshold} for more $\Omega$ values.)

Evaluating out-of-domain generalization with ANLI dataset splits resulted in an $\Omega_{\text{max}}$ value that is notably higher than $\mathcal{A}$ (89.7\% $\Omega_{\text{max}}$ for RoBERTa compared to 45.6\% $\mathcal{A}$). As a consequence, we encounter many \textit{flips}, i.e., examples where the model is unable to predict the gold label, but at least one permutation of that example is able to. However, recall this analysis expects us to know the gold label upfront, so this test can be thought of as running a word-order probe test on the model until the model predicts the gold label (or give up by exhausting our set of $q$ permutations). For out-of-domain generalization, $\Omega_{\text{rand}}$ decreases considerably (36.4\% $\Omega_{\text{rand}}$ on A1), which means fewer permutations are accepted by the model. Next, recall that a classic bag-of-words model would have $\mathcal{P}^c=100$ and $\mathcal{P}^f=0$. No model performs strictly like a classic bag of words although they do perform somewhat BOW-like ($\mathcal{P}^c >> \mathcal{P}^f$ for all test splits, \autoref{fig:comb_plot}). 
We find this BOW-likeness to be higher for certain non-Transformer models, (InferSent) as they exhibit higher $\mathcal{P}^c$ (84.2\% for InferSent compared to 70.7\% for RoBERTa on MNLI).

\begin{figure}[t]
    \centering
    \resizebox{0.48\textwidth}{!}{
        \includegraphics{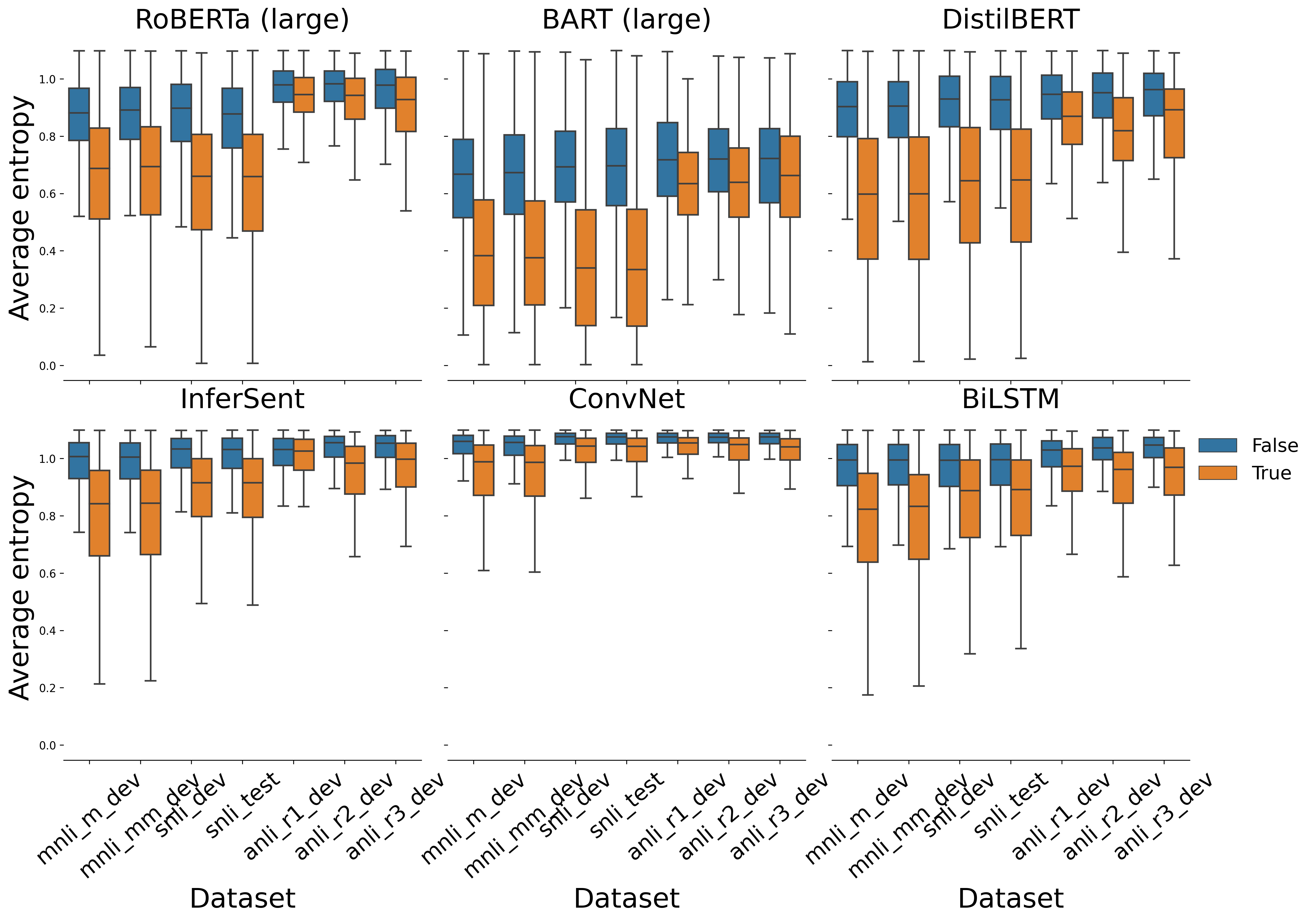}}
    \caption{Average entropy of model confidences on permutations that yielded the correct results for Transformer-based models (top) and Non-Transformer-based models (bottom). Results are shown for $D^c$ (orange) and $D^f$ (blue). The boxes show the quartiles of the entropy distributions.}
    \label{fig:all_entropy}
\end{figure}

\paragraph{Models are very confident.}

The phenomenon we observe would be of less concern if the correct label prediction was just an outcome of chance, which could occur when the entropy of the log probabilities of the model output is high (suggesting uniform probabilities on entailment, neutral and contradiction labels, recall Model B from \autoref{sec:permutation}). We first investigate the model probabilities for the Transformer-based models on the permutations that lead to the correct answer in \autoref{fig:all_entropy}. We find overwhelming evidence that model confidences on in-distribution datasets (MNLI, SNLI) are highly skewed, resulting in low entropy, and it varies among different model types. BART proves to be the most skewed Transformer-based model. This skewness is not a property of model capacity, as we observe DistilBERT log probabilities to have similar skewness as RoBERTa (large) model, while exhibiting lower $\mathcal{A}$, $\Omega_{\text{max}}$, and $\Omega_{\text{rand}}$. 



For non-Transformers whose accuracy $\mathcal{A}$ is lower, the $\Omega_{\text{max}}$ achieved by these models are also predictably lower. We observe roughly the same relative performance in the terms of $\Omega_{\text{max}}$ (\autoref{fig:comb_plot} and Appendix \autoref{table:main}) and Average entropy (Figure \ref{fig:all_entropy}). However, while comparing the averaged entropy of the model predictions, it is clear that there is some benefit to being a worse model---non-Transformer models are not as overconfident on randomized sentences as Transformers are. 
High confidence of Transformer models can be attributed to the \textit{overthinking} phenomenon commonly observed in deep neural networks \cite{kaya2019shallowdeep} and BERT-based models \cite{zhou2020bert}.

\paragraph{Similar artifacts in Chinese NLU.}

We extended the experiments to the Original Chinese NLI dataset \citep[OCNLI]{hu-etal-2020-ocnli}, and re-used the pre-trained RoBERTa-Large and InferSent (non-Transformer) models on OCNLI. Our findings are similar to the English results (\autoref{table:ocnli_all}), thereby suggesting that the phenomenon is not just an artifact of English text or tokenization.


\paragraph{Other Results.} We investigated the effect of sentence length (which correlates with number of possible permutations; \autoref{app_sec:length}), and hypothesis-only randomization (models exhibit similar phenomenon even when only hypothesis is permuted; \autoref{app_sec:HO}).

\begin{figure}[t]
    \centering
    \resizebox{0.48\textwidth}{!}{
        \includegraphics{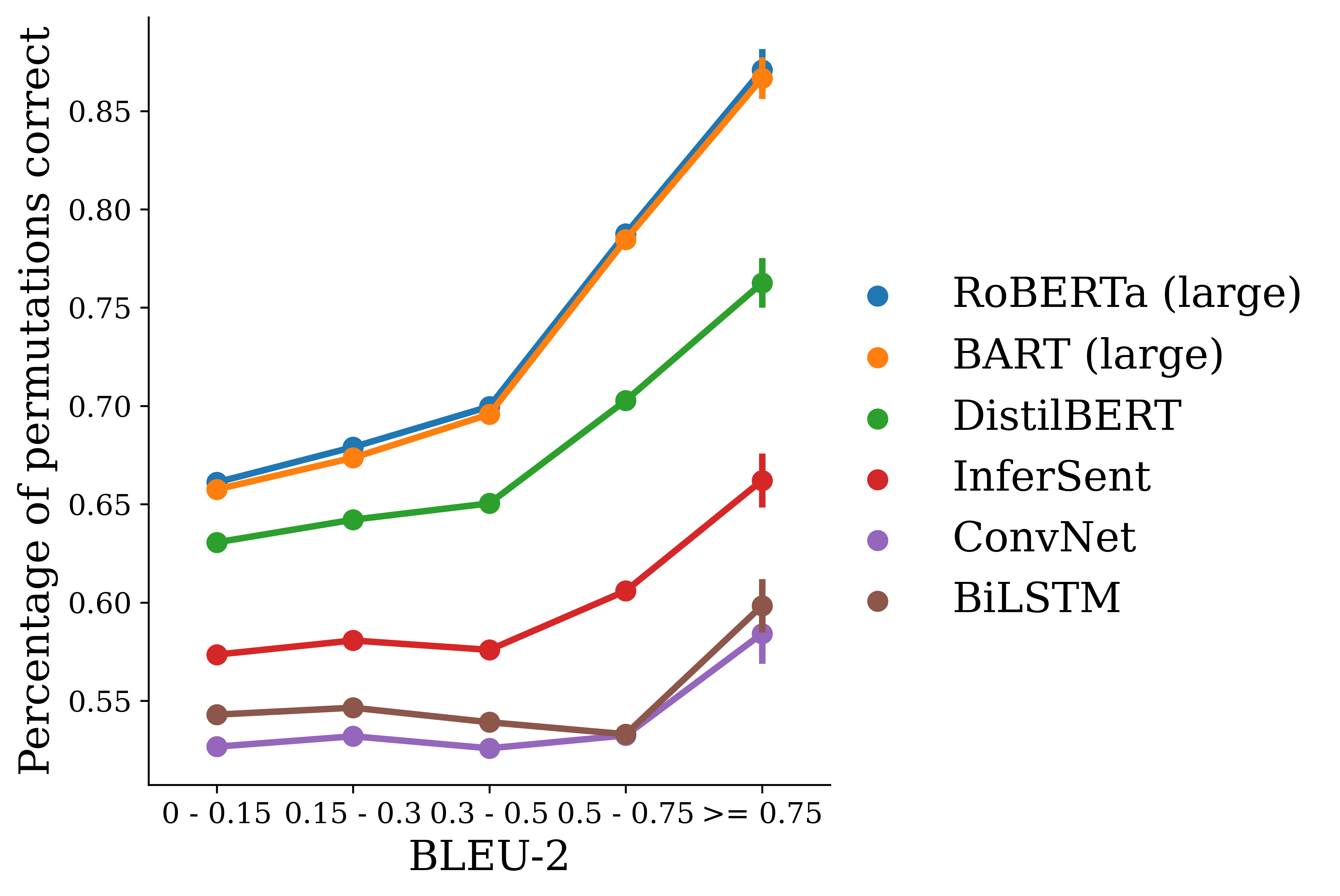}}
    \caption{BLEU-2 score versus acceptability of permuted sentences across all test datasets. RoBERTa and BART performance is similar but differs considerably from the performance of non-Transformer-based models, such as InferSent and ConvNet. }
    \label{fig:bleu_2}
\end{figure}

\section{Analyzing Syntactic Structure Associated with Tokens}\label{sec:pos_mini_tree}

A natural question to ask following our findings: what is it about particular permutations that leads models to accept them? Since the permutation operation is drastic and only rarely preserves local word relations, we first investigate whether there exists a relationship between \PermAcc\ scores and local word order preservation. Concretely, we compare bi-gram word overlap (BLEU-2) with the percentage of permutations that are deemed correct (\autoref{fig:bleu_2}).\footnote{We observe, due to our permutation process, the maximum BLEU-3 and BLEU-4 scores are negligibly low ($< 0.2$  BLEU-3 and $< 0.1$ BLEU-4), already calling into question the hypothesis that n-grams are the sole explanation for our finding. Because of this, we only compare BLEU-2 scores. Detailed experiments on specially constructed permutations that cover the entire range of BLEU-3 and BLEU-4 is provided in \autoref{app_sec:bleu_all}.} Although the probability of a permuted sentence to be predicted correctly does appear to track BLEU-2 score (Figure \ref{fig:bleu_2}), the percentage of examples which were assigned the gold label by the Transformer-based models is still higher than we would expect from permutations with lower BLEU-2 (66\% for the lowest BLEU-2 range of $0-0.15$), suggesting preserved relative word order alone cannot explain the high permutation acceptance rates.

Thus, we find that local order preservation does correlate with \PermAcc, but it doesn't fully explain the high \PermAcc\ scores.
We now further ask whether $\Omega$ is related to a more abstract measure of local word relations, i.e., part-of-speech (POS) neighborhood.

Many syntactic formalisms, like Lexical Functional Grammar \citep[LFG]{kaplan-bresnan-1995-formal, bresnan-etal-2015-lexical}, Head-drive Phrase Structure Grammar \citep[HPSG]{pollard-sag-1994-head} or Lexicalized Tree Adjoining Grammar \citep[LTAG]{schabes-etal-1988-parsing, abeille-1990-lexical}, are ``lexicalized'', i.e., individual words or morphemes bear syntactic features telling us which other words they can combine with. For example, ``buy'' could be associated with (at least) two lexicalized syntactic structures, one containing two noun phrases (as in \textit{\underline{Kim} bought \underline{cheese}}), and another with three (as in \textit{\underline{Lee} bought \underline{Logan} \underline{cheese}}). 
We speculate that our NLI models might accept permuted examples at high rates, 
because they are (perhaps noisily) reconstructing the original sentence from abstract, word-anchored information about common neighbors. 

To test this, we POS-tagged $D_{\text{train}}$ using 17 Universal Part-of-Speech tags (using spaCy, \citealt{spacy2}). 
For each $w_i \in S_i$, we compute the occurrence probability of POS tags on tokens in the \textit{neighborhood} of $w_i$. The neighborhood is specified by the radius $r$ (a symmetrical window $r$ tokens from $w_i \in S_i$ to the left and right). We denote this sentence level probability of neighbor POS tags for a word $w_i$ as $\psi^r_{\{w_i, S_i\}} \in \mathcal{R}^{17}$ (see an example in \autoref{fig:exPOSsignature} in the Appendix). Sentence-level word POS neighbor scores can be averaged across $ D_{\text{train}}$ to get a type level score $\psi^r_{\{w_i, D_{\text{train}}\}} \in \mathcal{R}^{17}, \forall w_i \in D_{\text{train}}$.  Then, for a sentence $S_i \in D_{\text{test}}$, for each word $w_i \in S_i$, we compute a \textbf{POS mini-tree overlap score}:
\begin{equation} 
\begin{split}
\beta^k_{\{w_i,S_i\}} =
\frac{1}{k} \mid \text{argmax}_k &\psi^r_{\{w_i, D_{\text{train}}\}} \cap \\ &\text{argmax}_k\psi^r_{\{w_i, S_i\}} \mid
\end{split}
\end{equation}

Concretely, $\beta^k_{\{w_i,S_i\}}$ computes the overlap of top-$k$ POS tags in the neighborhood of a word $w_i$ in $S$ with that of the train statistic. If a word has the same mini-tree in a given sentence as it has in the training set, then the overlap would be 1. For a given sentence $S_i$, the aggregate $\beta^k_{\{S_i\}}$ is defined by the average of the overlap scores of all its words: $\beta^k_{\{S_i\}} = \frac{1}{|S_i|}\sum_{w_i \in S_i} \beta^k_{\{w_i,S_i\}}$, and we call it a POS minitree \textit{signature}. We can also compute the POS minitree signature of a permuted sentence $\hat{S}_i$ to have $\beta^k_{\{\hat{S}_i\}}$.  If the permuted sentence POS signature comes close to that of the true sentence, then their ratio (i.e.,  $\beta^k_{\{\hat{S}_i\}} / \beta^k_{\{S_i\}}$) will be close to 1. Also, since POS signature is computed with respect to the train distribution, a ratio of $>$ 1 indicates that the permuted sentence is closer to the overall train statistic than to the original unpermuted sentence in terms of POS signature. 
If  high overlap with the training distribution correlates with percentage of permutations deemed correct, then our models treat words as if they project syntactic minitrees. 

\begin{figure}[t]
    \centering
    \resizebox{\linewidth}{!}{
        \includegraphics{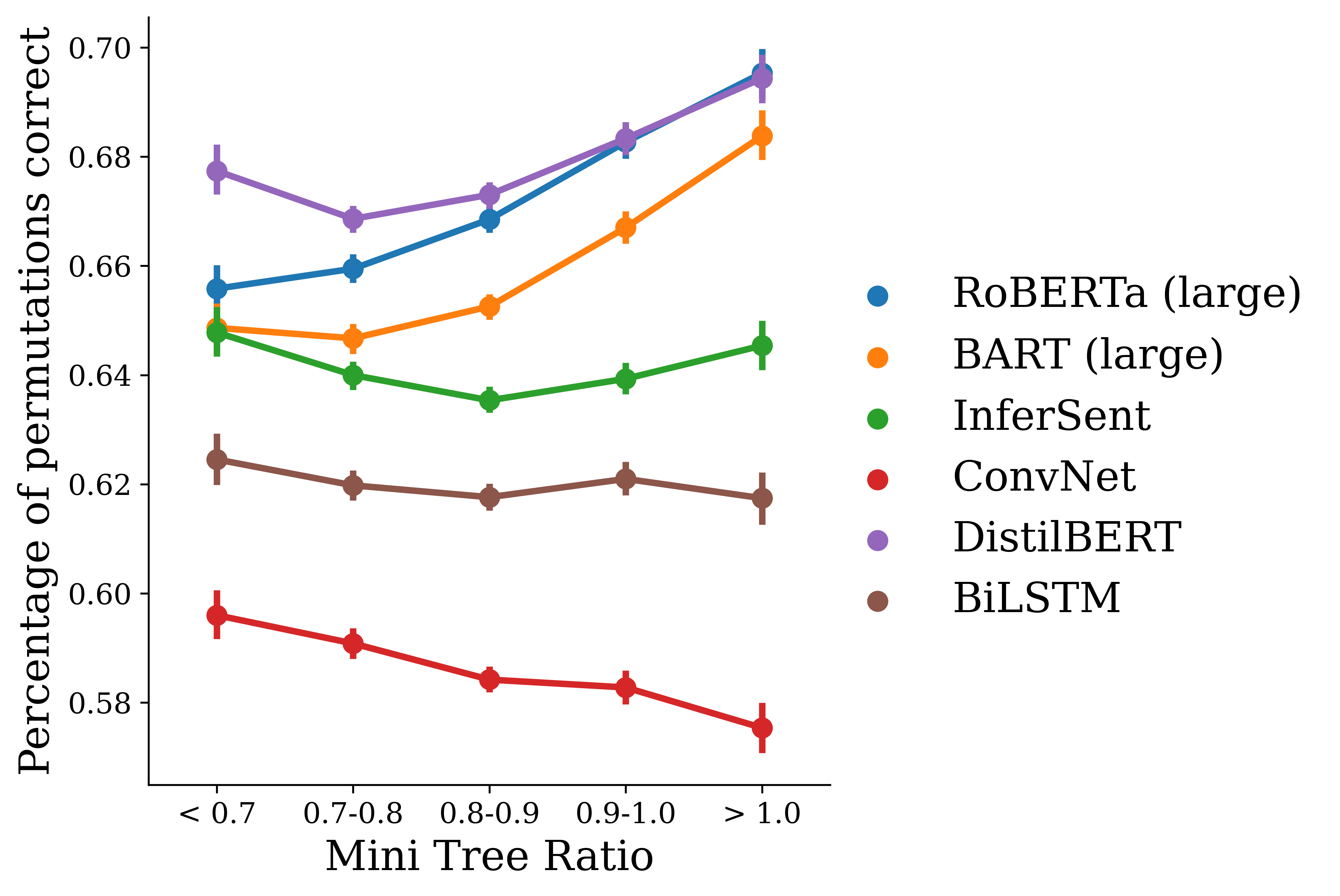}}
    \caption{POS Tag Mini Tree overlap score and  percentage of permutations which the models assigned the gold-label.}
    \label{fig:min_tree_4}
\end{figure}

We investigate the relationship with percentage of permuted sentences accepted with $\beta^k_{\{\hat{S}_i\}} / \beta^k_{\{S_i\}}$ in \autoref{fig:min_tree_4}. We observe that the POS Tag Minitree hypothesis holds for Transformer-based models, RoBERTa, BART and DistilBERT, where the percentage of accepted pairs increase as the sentences have higher overlap with the un-permuted sentence in terms of POS signature. For non-Transformer models such as InferSent, ConvNet, and BiLSTM models, the POS signature ratio to percentage of correct permutation remains the same or decreases, suggesting that the reasoning process employed by these models does not preserve local abstract syntax structure (i.e., POS neighbor relations).


\section{Human Evaluation}
\label{sec:human_eval}

We expect humans to struggle with UNLI, given our intuitions and the sentence superiority findings (but see \citealt{mollica-2020-composition}). To test this, we presented two experts in NLI (one a linguist) with permuted sentence pairs to label.\footnote{Concurrent work by \newcite{gupta-etal-2021-bert} found that untrained crowdworkers accept NLI examples that have been subjected to different kinds of perturbations at roughly most frequent class levels---i.e., only 35\% of the time.} Concretely, we draw equal number of examples from MNLI Matched dev set (100 examples where RoBERTa predicts the gold label, $D^c$ and 100 examples where it fails to do so, $D^f$), and then permute these examples using $\mathcal{F}$. The experts were given no additional information (recall that it is common knowledge that NLI is a roughly balanced 3-way classification task). Unbeknownst to the experts, all permuted sentences in the sample were actually accepted by the RoBERTa (large) model (trained on MNLI dataset). We observe that the experts performed much worse than RoBERTa (\autoref{tab:human_eval}), although their accuracy was a bit higher than random. We also find that for both experts, accuracy on permutations from $D^c$ was higher than on $D^f$, which verifies findings that showed high word overlap can give hints about the ground truth label \citep{dasgupta-etal-2018-evaluating, poliak-etal-2018-hypothesis, gururangan-etal-2018-annotation, naik-etal-2019-exploring}.

\begin{table}[t]
\centering
\resizebox{\linewidth}{!}{%
\begin{tabular}{@{}lllll@{}}
\toprule
Evaluator & Accuracy & Macro F1 & Acc on $D^{c}$ & Acc on $D^{f}$ \\ \midrule
X & 0.581 $\tiny\pm 0.068$ & 0.454 & 0.649 $\tiny\pm 0.102$ & 0.515 $\tiny\pm 0.089$ \\
Y &  0.378 $\tiny\pm 0.064$ & 0.378 & 0.411 $\tiny\pm 0.098$ & 0.349 $\tiny\pm 0.087$ \\ 
\bottomrule
\end{tabular}%
}
\caption{Human (expert) evaluation on 200 permuted examples from the MNLI matched development set. Half of the permuted pairs contained shorter sentences and the other, longer ones. 
All permuted examples were assigned the gold label by RoBERTa-Large. 
}
\label{tab:human_eval}
\end{table}

\section{Training by Maximizing Entropy}
\label{sec:training}

We propose an initial attempt to mitigate the effect of correct prediction on permuted examples. As we observe in \autoref{sec:eval}, model entropy on permuted examples is significantly lower than expected. 
Neural networks tend to output higher confidence than random for even unknown inputs \cite{gandhi2019mutual}, which might be an underlying cause of the high \PermAcc.


An ideal model would be ambivalent about randomized ungrammatical sentences. Thus, we train NLI models baking in the principle of mutual exclusivity \citep{gandhi2019mutual} by maximizing model entropy. Concretely, we fine-tune RoBERTa on MNLI while maximizing the entropy ($\bm{\mathrm{H}}$) on a subset of $n$ randomized examples ($(\hat{p}_i, \hat{r}_i$), for each example ($p,h$) in MNLI. We modify the loss function as follows:
\begin{dmath}
    \mathcal{L}=\argminB_{\theta}\sum_{\left((p, h),y\right)}y\log(p(y|(p,h);\theta)) + \sum_{i=1}^n \bm{\mathrm{H}}\left(y|(\hat{p}_i,\hat{h}_i);\theta\right)
\end{dmath}
Using this maximum entropy method ($n=1$), we find that the model improves considerably with respect to its robustness to randomized sentences, all while taking no hit to accuracy (\autoref{table:ME_train_roberta}). We observe that no model reaches a $\Omega_{\text{max}}$ score close to 0, suggesting further room to explore other methods for decreasing models' \PermAcc. Similar approaches have also proven useful \citep{gupta-etal-2021-bert} for other tasks as well.

\begin{table}[t]
\centering
\resizebox{\linewidth}{!}{%
\begin{tabular}{lrrrr}
\toprule
   Eval Dataset &   $\mathcal{A}$ (V) &  $\mathcal{A}$ (ME) &  $\Omega_{\text{max}}$ (V) &  $\Omega_{\text{max}}$ (ME) \\
\midrule
  MNLI\_m\_dev &  0.905 &     0.908 &    0.984 &        0.328 \\
 MNLI\_mm\_dev &  0.901 &     0.903 &    0.985 &        0.329 \\
   SNLI\_test &  0.882 &     0.888 &    0.983 &        0.329 \\
    SNLI\_dev &  0.879 &     0.887 &    0.984 &        0.333 \\
 ANLI\_r1\_dev &  0.456 &     0.470 &    0.890 &        0.333 \\
 ANLI\_r2\_dev &  0.271 &     0.258 &    0.880 &        0.333 \\
 ANLI\_r3\_dev &  0.268 &     0.243 &    0.892 &        0.334 \\
\bottomrule
\end{tabular}}
\caption{NLI Accuracy ($\mathcal{A}$) and \PermAcc\ metrics ($\Omega_{\text{max}}$) of RoBERTa when trained on MNLI dataset using vanilla (V) and Maximum Random Entropy (ME) method.}
  \label{table:ME_train_roberta}
\end{table}


\section{Future Work \& Conclusion}

We show that state-of-the-art models do not rely on sentence structure the way we think they should: NLI models (Transformer-based models, RNNs, and ConvNets) are largely insensitive to permutations of word order that corrupt the original syntax. 
We also show that reordering words can cause models to flip classification labels. 
We do find that models seem to have learned some syntactic information as is evidenced by a correlation between preservation of abstract POS neighborhood information and rate of acceptance by models, but these results do not discount the high rates of \PermAcc, and require further verification. Coupled with the finding that humans cannot perform UNLI at all well, the high rate of permutation acceptance that we observe leads us to conclude that current models do not yet ``know syntax'' in the fully systematic and humanlike way we would like them to.

A few years ago, \newcite{manning-etal-2015-computational} encouraged NLP to consider ``the details of human language, how it is learned, processed, and how it changes, rather than just chasing state-of-the-art numbers on a benchmark task.'' We expand upon this view, and suggest one particular future direction: we should train models not only to do well on clean test data, but also to not to overgeneralize to corrupted input. 

\section*{Acknowledgments}

Thanks to Omar Agha, Dzmitry Bahdanau, Sam Bowman, Hagen Blix, Ryan Cotterell, Emily Dinan, Michal Drozdal, Charlie Lovering, Nikita Nangia, Alicia Parrish, Grusha Prasad, Roy Schwartz, Shagun Sodhani, Anna Szabolsci, Alex Warstadt, Jackie Chi-kit Cheung, Timothy O'Donnell and members of McGill MCQLL lab for many invaluable comments and feedback on early drafts. 

\bibliography{acl2020}
\bibliographystyle{acl_natbib}

\clearpage

\appendix


\begin{figure}[t]
    \centering
    \resizebox{\linewidth}{!}{
        \includegraphics{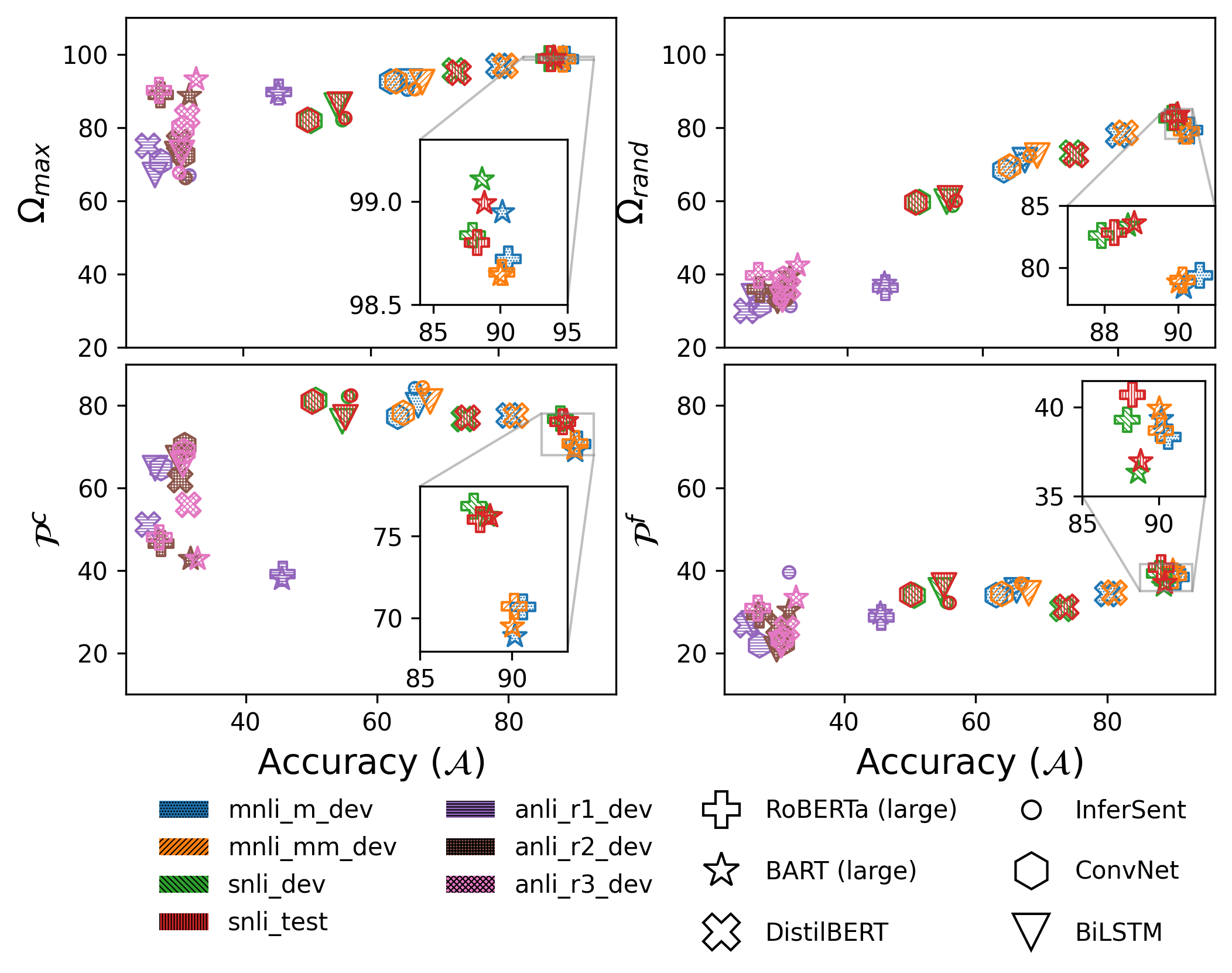}}
    \caption{Comparison of $\Omega_{\text{max}}$,$\Omega_{\text{rand}}$,$\mathcal{P}^c$ and $\mathcal{P}^f$ with the model accuracy $\mathcal{A}$ on multiple datasets, where all models are trained on the MNLI corpus \cite{williams-etal-2018-broad}.}
    \label{fig:comb_plot}
\end{figure}



\section{Effect of Length on \PermAcc}
\label{app_sec:length}

\begin{figure}[ht]
    \centering
    \resizebox{0.5\textwidth}{!}{
        \includegraphics{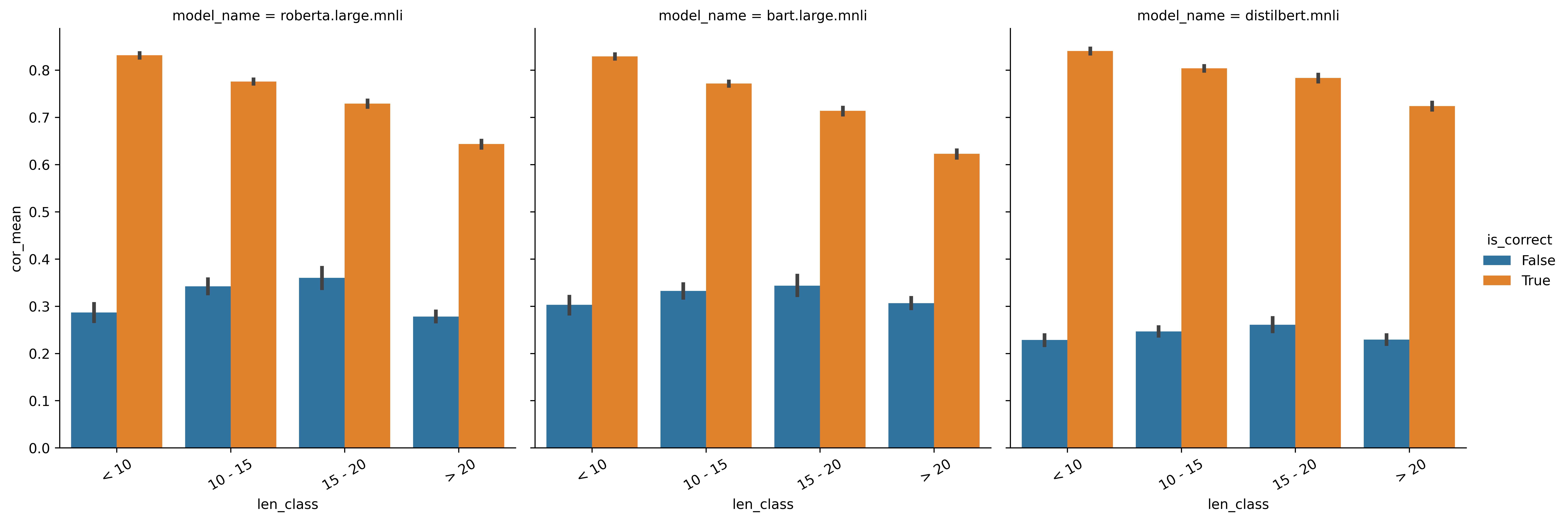}}
    \caption{Length and \PermAcc by Transformer-based models.}
    \label{fig:length}
\end{figure}

We investigate the effect of length on \PermAcc\ in Figure \ref{fig:length}. We observe that shorter sentences in general have a somewhat higher probability of acceptance for examples which was originally predicted correctly---since shorter sentences have fewer unique permutations. However, for the examples which were originally incorrect, the trend is not present.  

\section{Example of POS Minitree}

\begin{figure}[ht]
    \centering
    \resizebox{0.5\linewidth}{!}{
        \includegraphics{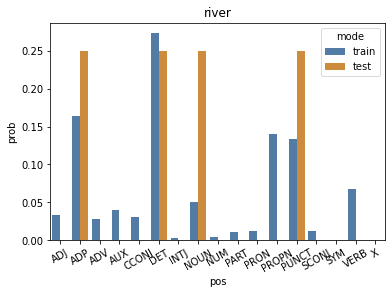}}
    \caption{Example POS signature for the word `river', calculated with a radius of 2. Probability of each neighbor POS tag is provided. Orange examples come from the permuted test set, and blue come from the original training data. }
    \label{fig:exPOSsignature}
\end{figure}

In \autoref{sec:pos_mini_tree}, we developed a POS signature for each word in at least one example in a test set, then compare that signature to the distribution of the same word in the training set. \autoref{fig:exPOSsignature} provides a snapshot a word ``river" from the test set and shows how the POS signature distribution of the word in a particular example match with that of aggregated training statistic. In practice, we select the top $k$ POS tags for the word in the test signature as well as the train, and calculate their overlap. When comparing the model performance with permuted sentences, we compute a ratio between the original test overlap score and an overlap score calculated instead from the permuted test. In the \autoref{fig:exPOSsignature}, `river' would have a POS tag minitree score of 0.75.

\begin{figure*}
    \centering
    \resizebox{\textwidth}{!}{
        \includegraphics{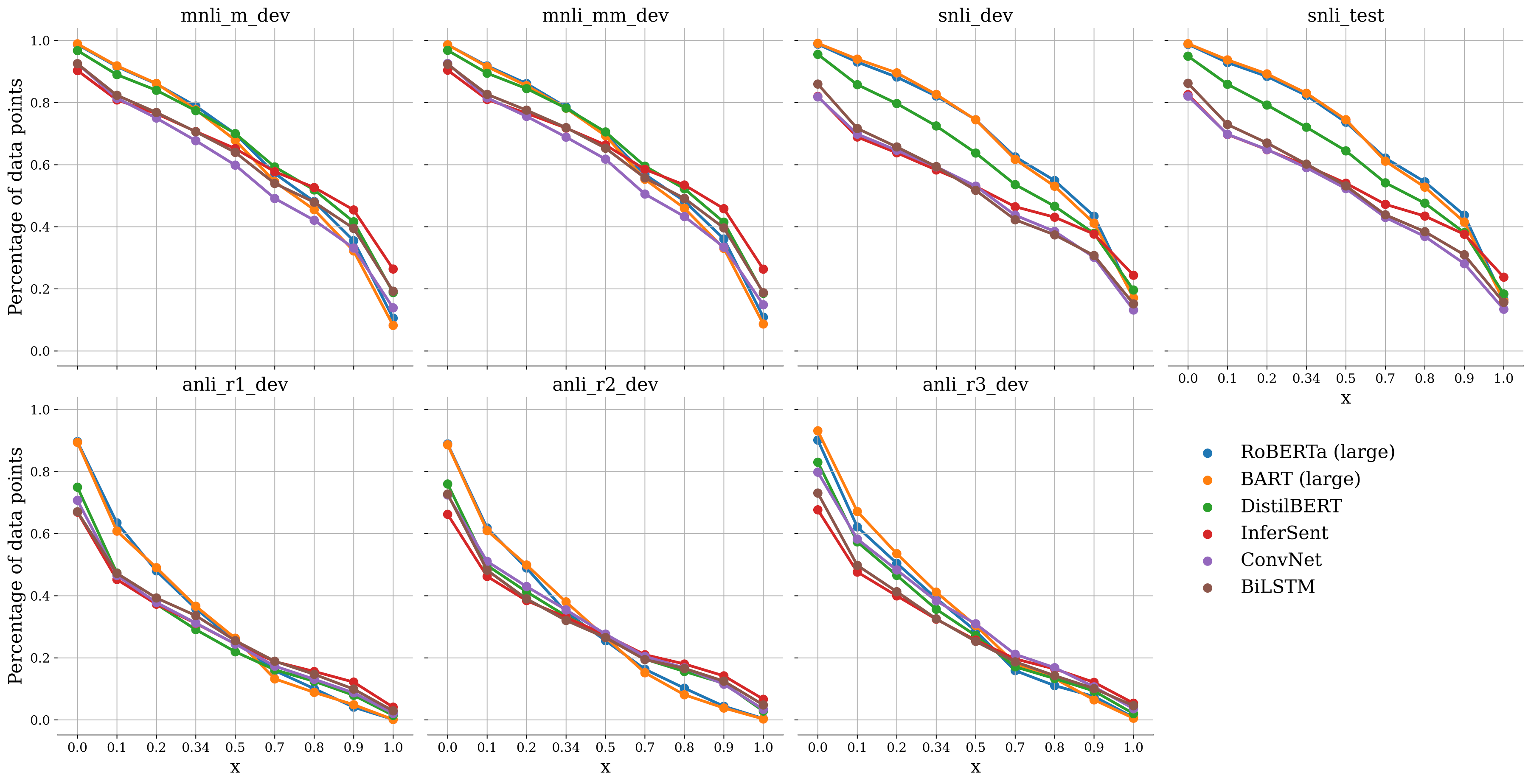}}
    \caption{$\Omega_x$ threshold for all datasets with varying $x$ and computing the percentage of examples that fall within the threshold. The top row consists of in-distribution datasets (MNLI, SNLI) and the bottom row contains out-of-distribution datasets (ANLI)}
    \label{fig:threshold_omega_x}
\end{figure*}

\section{Effect of Hypothesis-Only Randomization}
\label{app_sec:HO}

\begin{figure*}[ht]
    \centering
    \subfigure[]{
    \resizebox{0.48\textwidth}{!}{
        \includegraphics{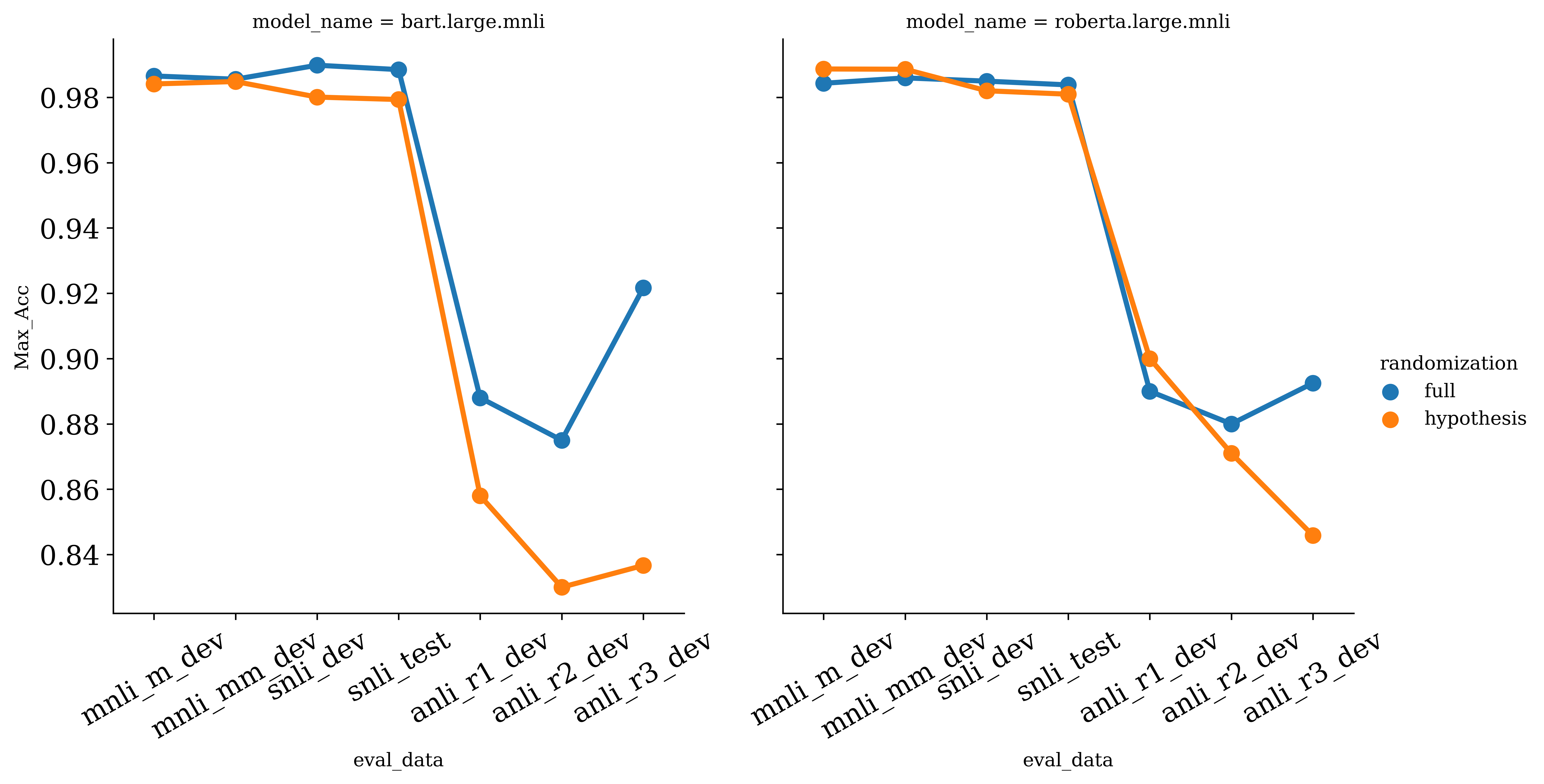}}
    \label{subfig:compare}
    }
    \subfigure[]{
    \resizebox{0.48\textwidth}{!}{
        \includegraphics{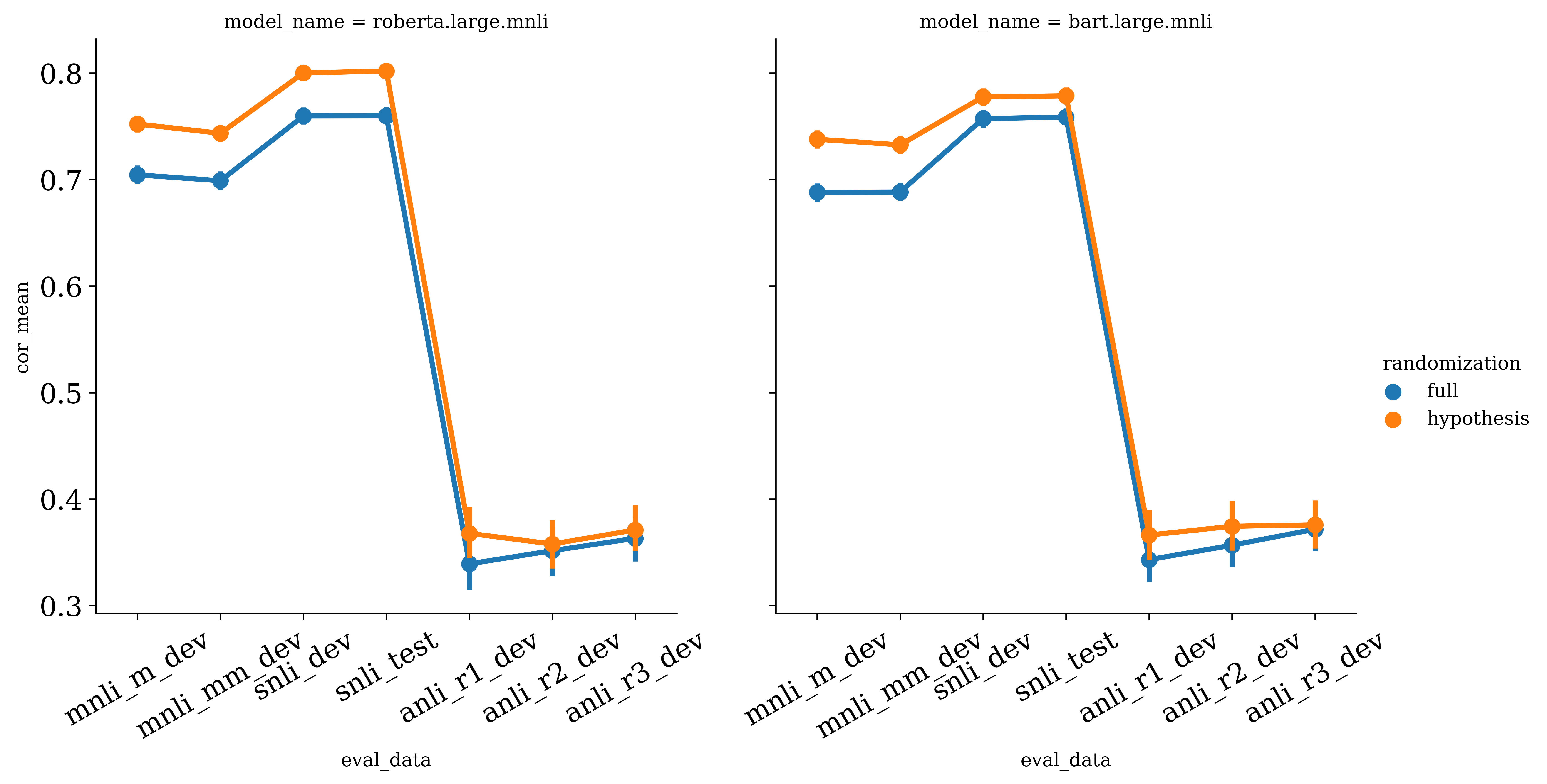}}
    \label{subfig:compare_cor}
    }
    \caption{Comparing the effect between randomizing both premise and hypothesis and only hypothesis on two Transformer-based models, RoBERTa and BART (For more comparisons please refer to Appendix). In \ref{subfig:compare}, we observe the difference of $\Omega_{\text{max}}$ is marginal in in-distribution datasets (SNLI, MNLI), while hypothesis-only randomization is worse for out-of-distribution datasets (ANLI). In \ref{subfig:compare_cor}, we compare the mean number of permutations which elicited correct response, and naturally the hypothesis-only randomization causes more percentage of randomizations to be correct.}
    \label{fig:hypothesis_compare}
\end{figure*}

In recent years, the impact of the hypothesis sentence \citep{gururangan-etal-2018-annotation, tsuchiya-2018-performance, poliak-etal-2018-hypothesis} on NLI classification has been a topic of much interest. As we define in \autoref{sec:permutation}, logical entailment can only be defined for pairs of propositions. We investigated one effect where we randomize only the hypothesis sentences while keeping the premise intact. Figure \ref{subfig:compare} shows that the $\Omega_{\text{max}}$ value is almost the same for the two schemes; randomizing the hypothesis alone also leads the model to accept many permutations.

\section{Effect of clumped words in random permutations}
\label{app_sec:bleu_all}

\begin{figure}
    \centering
    \resizebox{0.48\textwidth}{!}{
        \includegraphics{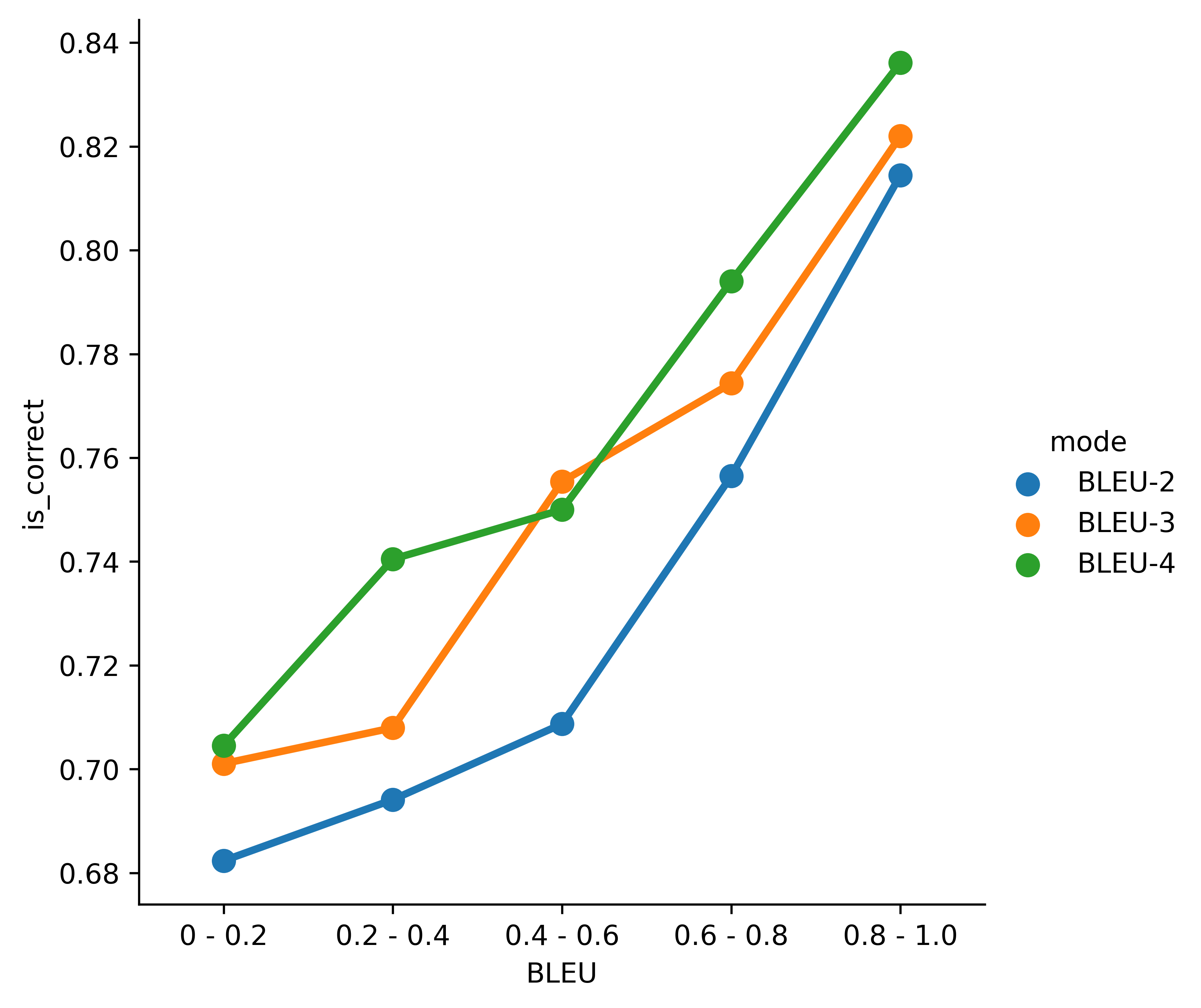}}
    \caption{Relation of BLEU-2/3/4 scores against the acceptability of clumped-permuted sentences accross all test datasets on all models. }
    \label{fig:bleu_2_3_4}
\end{figure}

Since our original permuted dataset consists of extremely randomized words, we observe very low BLEU-3 ($<$ 0.2) and BLEU-4 scores ($<$ 0.1). To study the effect of overlap across a wider range of permutations, we devised an experiment where we clump certain words together before performing random permutations. Concretely, we clump 25\%, 50\% and 75\% of the words in a sentence and then permute the remaining words and the clumped word as a whole. This type of clumped-permutation allows us to study the full range of BLEU-2/3/4 scores, which we present in \autoref{fig:bleu_2_3_4}. As expected, the acceptability of permuted sentences increase linearly with BLEU score overlap. 

\section{Effect of the threshold of $\Omega_x$ in various test splits}
\label{app_sec:threshold}

We defined two variations of $\Omega_x$, $\Omega_{\text{max}}$ and $\Omega_{\text{rand}}$, but theoretically it is possible to define any arbitrary threshold percentage $x$ to evaluate the unnatural language inference mechanisms of different models. In \autoref{fig:threshold_omega_x} we show the effect of different thresholds, including $\Omega_{\text{max}}$ where $x = 1/|D_{\text{test}|}$ and $\Omega_{\text{rand}}$ where $x = 0.34$. We observe for in-distribution datasets (top row, MNLI and SNLI splits), in the extreme setting when $x=1.0$, there are more than 10\% of examples available, and more than 25\% in case of InferSent and DistilBERT. For out-of-distribution datasets (bottom row, ANLI splits) we observe a much lower trend, suggesting generalization itself is the bottleneck in permuted sentence understanding.




\section{Training with permuted examples}

In this section, we hypothesize that if the NLU models are mostly insensitive to word order, then training using permuted examples should also yield the same or comparable accuracy as training using grammatically correct data (i.e., the \textit{standard setup}). To test this, we train Transformer-based models on top of $\hat{D}_{\text{train}}$, which is computed by applying $\mathcal{F}$ on each example of $D_{\text{train}}$ for $q=1$ times. This ensures a control case where we keep the same amount of training data as the standard setup (such that models does not benefit from data augmentation). We also ensure that we use the same hyperparameters while training as with the standard setup. Concretely, $\hat{D}_{\text{train}}$ consists of $n$ hypothesis-premise pairs from MNLI training data, where each example is a permuted output of the original pair. 

\begin{table}
\centering
\resizebox{\linewidth}{!}{%
\begin{tabular}{lcccccc}
\toprule
Eval Data & \multicolumn{2}{c}{RoBERTa} & \multicolumn{2}{c}{BART} & \multicolumn{2}{c}{DistilBERT} \\
 & $\mathcal{A}$ & $\hat{\mathcal{A}}$ & $\mathcal{A}$ & $\hat{\mathcal{A}}$ & $\mathcal{A}$ & $\hat{\mathcal{A}}$ \\
\midrule
MNLI Matched & 0.906 & 0.877 & 0.902 & 0.862 & 0.800 & 0.760 \\
MNLI Mismatched & 0.901 & 0.878 & 0.900 & 0.869 & 0.811 & 0.769 \\
SNLI Dev & 0.879 & 0.870 & 0.886 & 0.854 & 0.732 & 0.719 \\
SNLI Test & 0.883 & 0.873 & 0.888 & 0.859 & 0.738 & 0.719  \\
ANLI R1 (Dev) & 0.456 & 0.367 & 0.455 & 0.336 & 0.251 & 0.250 \\
ANLI R2 (Dev) & 0.271 & 0.279 & 0.316 & 0.293 & 0.300 & 0.290  \\
ANLI R3 (Dev) & 0.268 & 0.271 & 0.327 & 0.309 & 0.312 & 0.312 \\
\bottomrule
\end{tabular}}
 \caption{Statistics for Transformer-based models when trained on permuted MNLI corpus. We compare the accuracy for both models trained on unpermuted data ($\mathcal{A}$) and the permuted data ($\hat{\mathcal{A}}$). We use original test sets during inference.}
\label{tab:train_with_rand}
\end{table}

We present the results of such training in \autoref{tab:train_with_rand}, and compare the accuracy ($\hat{\mathcal{A}}$) with that of the standard setup ($\mathcal{A}$). Note, during inference for all the models we use the un-permuted examples.
As we can see, models perform surprisingly close to the original accuracy $\mathcal{A}$ \textit{even when trained with ungrammatical sentences}. This adds further proof to the BOW nature of NLU models.



\begin{table}
\centering
\resizebox{\linewidth}{!}{%
\begin{tabular}{lrr}
\toprule
Dataset & Test Examples & Used Examples\\
\midrule
MNLI Matched & 9815 & 6655\\
MNLI Mismatched & 9832 & 7449\\
SNLI Dev & 9842 & 3697\\
SNLI Test & 9824 & 3671\\
ANLI R1 (Dev) & 1000 & 756\\
ANLI R2 (Dev) & 1000 & 709\\
ANLI R3 (Dev) & 1200 & 754\\
\bottomrule
\end{tabular}}
 \caption{Dataset statistics used in this paper for inference. `Used Examples' provides the number of premise-hypothesis pairs for the dataset which we selected for inference (i.e., examples where at least 100 unique permutations were possible).}
\label{tab:data_stats}
\end{table}


\section{Reproducibility Checklist}

As per the prescribed Reproducibility Checklist, we provide the information of the following:

\begin{itemize}
    \item \textit{A clear description of the mathematical setting, algorithm and/or model}: We provide details of models used in \autoref{sec:eval}
    \item \textit{Description of the computing infrastructure used}: We used 8 NVIDIA V100 32GB GPUs to train the models and perform all necessary inferences. We didn't run hyperparameter tuning for Transformer-based models as we used the published hyperparameters from the original models.
    \item \textit{Average runtime for each approach}: On an average, each model inference experiment consistine of 100 permutations for each example takes roughly 1 hour to complete.
    \item \textit{Relevant statistics of the datasets used}: We provide the statistics of the datasets used in \autoref{tab:data_stats}.
    \item \textit{Explanation of any data that were excluded, and all pre-processing steps}: We exclude examples where either the hypothesis and premise consists of less than 6 tokens. This way, we ensure that we have 100 unique permutations for each example.
    \item \textit{Link to downloadable version of data and code}: We provide downloadable version of our data and code at \href{https://github.com/facebookresearch/unlu}{https://github.com/facebookresearch/unlu}.
\end{itemize}

\end{document}